# Creating a contemporary corpus of similes in Serbian by using natural language processing

Nikola Milošević and Goran Nenadić

School of Computer Science, University of Manchester

Nikola.milosevic@manchester.ac.uk, g.nenadic@manchester.ac.uk

*Abstract*
Simile is a figure of speech that compares two things through the use of connection words, but where comparison is not intended to be taken literally. They are often used in everyday communication, but they are also a part of linguistic cultural heritage. In this paper we present a methodology for semi-automated collection of similes from the World Wide Web using text mining techniques. We expended an existing corpus by collecting 442 similes from the internet and adding it to the existing Karadžić's corpus that contained 333. We, also, introduce crowdsourcing to the collection of figures of speech, which helped us build corpus containing 787 unique similes

## 1. Introduction

Similes in Serbian language are common figures of speech. Simile is a figure of speech which compares two things through the use of connection words (Harris and others, 2002), but where comparison is not intended to be taken literally. Similes are metaphors with the only difference that they use words such as "like" and "as". In other words, a simile makes the comparison explicit (Cooper, 1986; Niculae, 2013). As it is usual in figures of speech, simile are figurative and in many cases amusing. They are used commonly in both written and spoken language. Vuk Karadžić first attempted to collect similes, together with other short figures of speech (Karadžić, 1849). Some of the similes Karadžić collected include examples such as "bela kao sneg" (white like a snow), "crveni se kao oderano goveče" (red as skinned ox), "gladan kao kurjak" (hungry like a wolf), etc. Together with other figures of speech, such as proverbs and puzzles, similes in Serbian language represent an unique kind of linguistic cultural heritage.

"Language is a living thing. We can feel it changing. Parts of it become old: they drop off and are forgotten. New pieces bud out, spread into leaves, and become big branches, proliferating" noticed Gilbert Highet (Highet, 1971). Changes in language affect also short figures of speech. Some of the phrases Vuk Karadžić collected are no more in use, while new emerged. There is an emerging need for updating corpora of various figures of speech.

World Wide Web provided people a novel ways to express themselves on a variety of media such as Wikis, question-answer databases, blogs, forums, review-sites and social media. Estimation is that around 50 million scholarly articles were published in our history (Jinha, 2010). Google estimated that there are more than 129 million published books in the World (Taycher, 2010), while there are about 4.6 billion web pages in the indexed web and nearly 550 billion individual documents in non-indexed web (Bergman, 2001). Projections say that the amount of data generated on the web will increase by 40% annually (Larose, 2014). The amount of content that has been published on the World Wide Web is probably larger than the amount of published literature on the paper in history. World Wide Web became a great resource for exploring changes and trends in language, especially in the written form. People are using modern language on the World Wide Web and by analysing it, we aimed to collect new similes and other common short forms.

In this paper we present our methodology for creating a corpus of similes in Serbian language from the World Wide Web by using natural language processing and machine learning techniques. Our methodology is semi-automatic and involves a curator who checks and approves the data. We have applied our method to the few popular forums and websites, which led to the creation of unique corpus of simile used in modern Serbian language.

## 2. Background

Vuk Stefanović Karadžić made a great collection of phrases (Karadžić, 1852), similes, proverbs (Karadžić, 1849), puzzles (Karadžić, 1897), folk poems (Karadžić, 1977; Karadžić, 1976; Karadžić, 1814), folk stories (Karadžić, 1953) and customs (Karadžić, 1867). During the 19[th] century there were, also a number of magazines that were publishing folk poems, stories and phrases, such as Danica (Karanović, 1990), Vila (Maticki, 1985) and Letopis Matice srpske (Maticki & Jerković, 1983; Samardžija, 1995). By the end of 19[th] century most of these magazines vanished or stopped collecting folks' art and literature. During the 20[th] century, there were some attempts to collect proverbs and other short forms from some particular geographical area (Marković, 1979; Jevtić, 1969; Vukanović, 1983; Cvetanović, 1980). Vukajlija website emerged in 2007. as a crowd-sourced dictionary of Serbian slang[1] , which collected and defined some of the modern phrases and other figures of speech. However, most of the interest in 20[th] and 21[th] century was on proverbs and little has been done in updating similes and other figure of speech corpora in Serbian.

In English, Veale and Hao (Veale & Hao, 2007) extracted explicit similes they modelled as "X is as P as Y", where X and Y are nouns and P is an Adjective. They used Google search API to query the internet and collect similes in the modelled form. Li et al. (Li, et al., 2012) used similar approach querying Baidu search engine for Chinese similes to extract set of pairs that are compared. They used three models for similes in Chinese.

---
[1] http://vukajlija.com

There are multiple types of similes. A study of similes in English and Norwegian takes into account two types: *like a/an + noun* and *as + adjective + as + noun* (Aasheim, 2012). In Chinese were studied three simile templates in text mining research, which translated to English would look like *as + noun+ same*, *as + verb + same* and *as + same + adjective* (Li, et al., 2012).

In Slavic languages, simile expressions are based on the preposition "like" or "as" (*kak* in Russian, *kao* in Serbian, *ako* in Slovak, etc.) and introduce a noun phrase (NP) agreeing in case with the standard of comparison. When such a construction is applied to a possessive pronoun, the complement of the preposition is a noun phrase in the genitive case (Rappaport, 1999).

To the best of our knowledge, there was no text mining approach to collecting similes in Serbian. Rare are collecting approaches for similes in other Slavic languages. We developed a semi-automatic approach to mining similes in Serbian from the web.

## 3. Method

### 3.1 Method overview

Our methodology for collecting similes in Serbian consists of four steps: firstly we collect the documents for processing, in the second step we process these documents and extract all comparisons, in the third step we use machine learning to distinguish comparisons that are similes from the others, and in the final step, human curator reviews the data and corrects the mistakes. Also, users and visitors of our website can propose new similes and help that way to collect the similes that were not collected by our text mining pipeline. Curator needs to check these manually imputed similes. Workflow of our methodology can be seen in the *Figure 1*: Overview of proposed methodology consisting of four steps: 1) Document retrieval using a web crawler, 2) comparison extraction which is using part-of-speech tagging, 3) classification of similes and 4) data curation which require human curator to check.

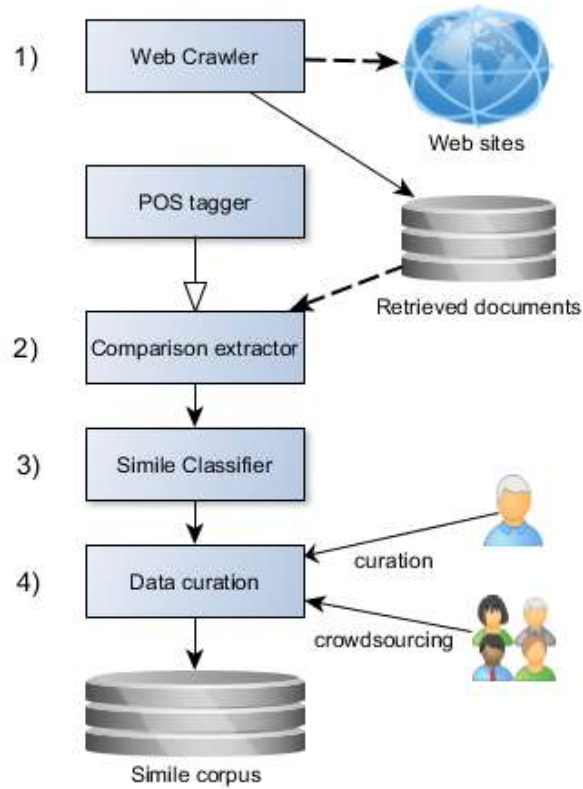

*Figure 1: Overview of proposed methodology consisting of four steps: 1) Document retrieval using a web crawler, 2) comparison extraction which is using part-of-speech tagging, 3) classification of similes and 4) data curation which require human curator to check*

We have, also created a model that is able to computationally match the similes in text.

### 3.2 Modelling similes in Serbian

In Serbian, we can identify two main categories of similes, which can be modelled in the following fashion:

- Adjective + Connection word (kao - like, as) + Noun Phrase (i.e. "lep kao cvet" – "beautiful like a flower")

- Verb + Connection word (kao - like, as) + Noun Phrase (i.e. "radi kao konj" – "works like a horse")

Generally, this could be written in the following way:

$$(V|A|V\ se)\ (kao|ko|k'o)\ (NP) \qquad (1)$$

A typical noun phrase consists of a noun together with zero or more dependents of various types, such as determiners, adjectives, adjective phrases, noun adjuncts, prepositional phrases, participle phrases or pronouns (Crystal, 2011). Connection word used in similes in Serbian is "*kao*", but it, also could be found as a shorter form of a word like "*k'o*" or "*ko*".

Currently available part-of-speech taggers for Serbian can tag only words, but not phrases. Because of this, it was not possible to rely on tagger to identify noun phrases.

We modelled similes as a verb or adjective, followed by explicit use of connection words "kao", "k'o" or "ko", followed by one or more adjectives and terminated with a noun (see Model 2)

$$(V|A|V\backslash\ se)\ (kao|ko|k'o)\ \ (A^*)\ (N) \qquad (2)$$

Noun phrase model of *(A\*) (N)* is modelling majority of commonly used noun phrases in Serbian. However, there can be noun phrases and similes that are not terminated with the first noun. Example could be "smoren kao zmaj u vatrogasnoj stanici" (translation:"bored like a dragon at a fire station"). In this example word "zmaj" which means "dragon" is a first noun that would terminate simile. However, "smoren kao zmaj" is a commonly used simile, so our method would not make mistake totally, but it will miss the extended version of the simile. However, these extended versions are not frequent and we will leave them for manual curation in the later stage of our pipeline.

Our model is extracting comparisons with similes quite well, however, when we looked at the results, we found that it is generating a large amount of false positives. Simile model will find simile such as "radi kao konj" (translation: "works like a horse"), but it will also extract phrases like "radi kao pravnik" (translation: "works as a lawyer"). The first phrase is a simile, while the second is not. There is no lexical feature that would give a cue that these two phrases are in any way different. Only semantics differentiate similes. Current tools for Serbian language do not allow us to efficiently make separations between phrases on semantic level. It turned out that only about 10% of extracted phrases were true similes. In the first instance we relied on curator's manual review of the phrases.

### 3.3 Crawler

In order to collect textual data in Serbian language, we developed a number of crawlers crafted for the particular websites. We wanted to download all meaningful text from these websites, but skip contentless parts of the website, such as menus, headings and footers. In order to do that, each crawler was extracting only text that is inside some div HTML tag with particular id. Crawlers were following links on each page they visit, but not outside the domain of the website they were crafted for. Particularly, we developed four crawlers that were extracting text from the four different websites:

- *laguna.rs* - one of the biggest book publisher with abstracts, reviews and parts of the books on their website;

- *rastko.rs* - website of a project that is trying to make a digital library of books and articles in Serbian language that are considered cultural heritage;

- *burek.com* - a large general public forum on Serbian;

- *tarzanija.com* - a popular blog portal with sarcastic comments on various issues.

Texts extracted from these websites were saved as text files. For crawling and processing we used Scrapy - an open source Python framework for extracting data from website[2].

### 3.4 Simile extractor

In the next step of our pipeline we are using extracted textual documents to extract comparisons. Since our model of similes requires part of speech tagging, we used a part of speech tagger for Croatian and Serbian language (Agić, et al., 2013) . This tagger is publicly available as a POS tag model for HunPos tagger (Halacsy, et al., 2007). We used model 2 for matching similes.

### 3.5 Simile classifier

While reviewing first set of outputted phrases, curator has created decent amount of true positive and false positive examples that could be used for machine learning. We created a machine-learning based classifier that can distinguish true simile from the other phrases that have same lexical characteristics.

For machine learning we used Weka toolkit (Hall, et al., 2009). We used 300 examples of true positive similes and 300 examples of false positive in order to train and test our algorithm. As a features for our machine learning algorithm we used whole phrase (whole simile, for illustration we will use "radi kao konj", which can be translated to English as "works as a horse"), stemmed phrase ("rad ka konj"), left side of phrase that is before connection word ("radi"), left side stemmed ("rad"), part of phrase that is on the right from the connection word ("konj") and stemmed right part ("konj"). We tested learning algorithms with various combinations of these features by dropping some of them. For stemming we used stemmer for Serbian language (Milošević, 2012), that we ported for Python[3].

---

[2] http://scrapy.org/
[3] https://github.com/nikolamilosevic86/SerbianStemmer

As a machine learning algorithm we experimented with Multinominal Naive Bayes, Random Forests and Support Vector Machines (SVM) with polynomial kernel that uses sequential minimal optimization. The models exported from Weka are downloadable from our website[4].

### 3.6 Curation interface

When similes were extracted from the text, they were added to mySQL database. We created a website for viewing, searching and editing our dataset that is using this database. This website has the following curation features:

- **Viewing of currently curated similes**, sorted alphabetically (See *Figure 2*:Interface for viewing collecting similes sorted alphabetically)

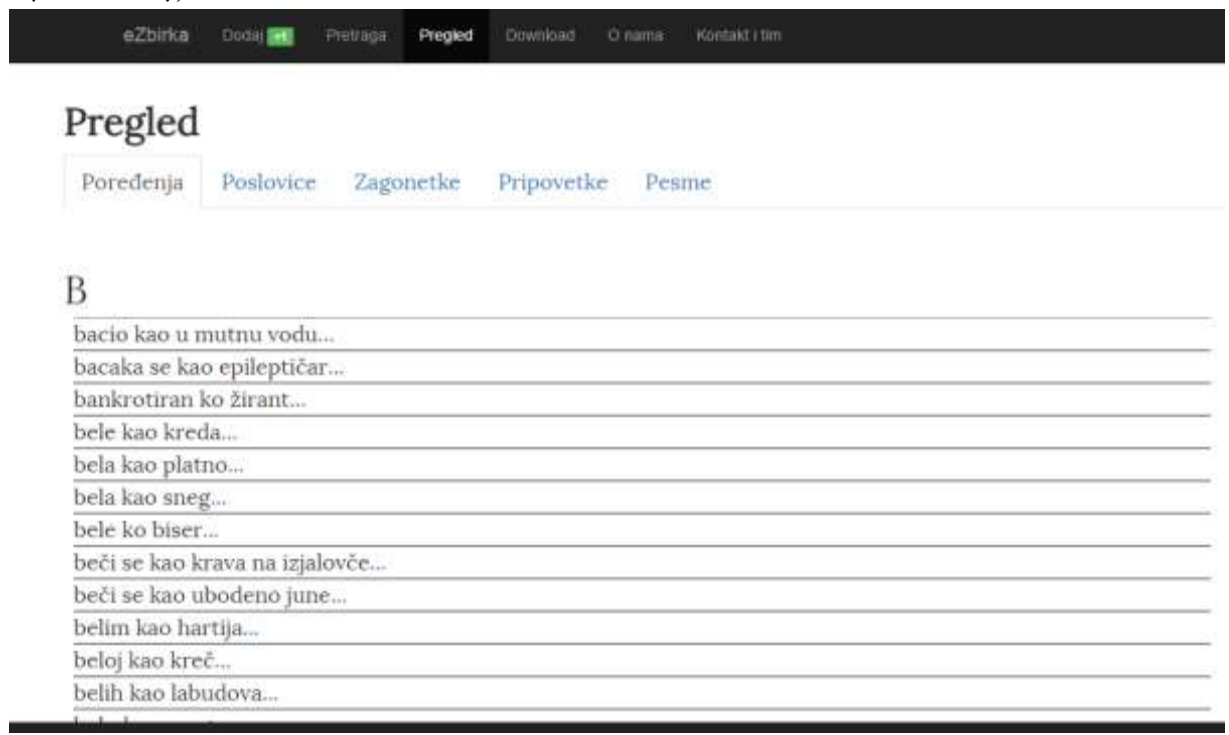

*Figure 2:Interface for viewing collecting similes sorted alphabetically*

- **Searching for simile**. Searching is using the stemmer for Serbian, so it can find similes that are not in exactly the same form in the database as in the query. Since Serbian language contains a lot of inflections and we tend to store only one instance (which also means one kind of inflection) of simile, stemming is useful to find the queried simile even if stored in different word form. For example "beo kao sneg" (m), "bela kao sneg" (f) and "belo kao sneg" (n)

---

[4] http://ezbirka.starisloveni.com/download.html

(translation: "white as a snow" in three different grammatical genders - masculine, feminine, neuter), will be treated by our search algorithm as a same set of words. Search interface is shown in *Figure 3*:Interface for searching collecting similes.

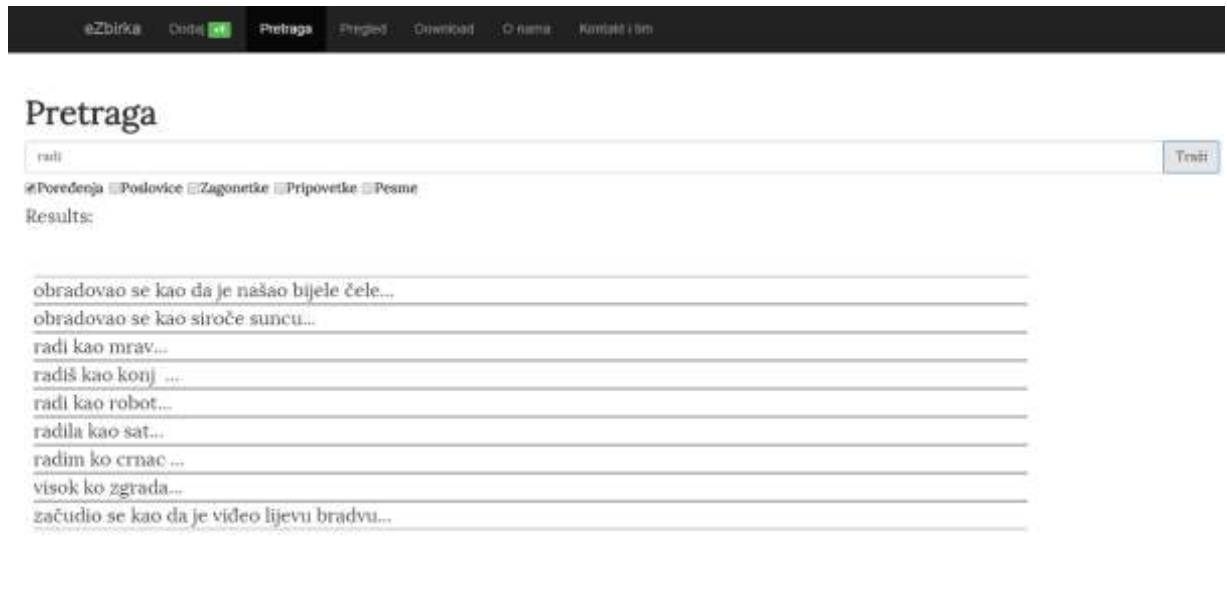

*Figure 3:Interface for searching collecting similes*

- **Adding a new simile to the database**. Since some of the similes might be rare or just not used in the corpus we were examining, users of the website are able to add manually similes that are missing. If the user wants to add simile that already exists in the database in some form, user will be notified. This is done using stemming process for both added simile and the similes already stored in the database. The user will be presented with similar similes in database and can decide whether to add it anyway. However, the simile won't be visible on the public website until curator approves it. Interface is presented in *Figure 4*: Interface for adding new figure of speech into the collection.

*Figure 4: Interface for adding new figure of speech into the collection*

- **Curation interface**. User can login to this part of the website and perform various curation tasks over the similes. On the first screen, after login, user will be presented with similes that are still not approved. User can approve them, reject them (which will delete it from the list, but not from the database) or edit them.

## 4. Results

Corpora we obtained from the web contained 40239 documents (more detailed information about the dataset can be seen in *Table 1*. Number of documents in dataset obtained by crawlers.). Using our semi-automated workflow we extracted potential similes and manually reviewed them. Manual review of a data reviled that among potential similes we had around 5000 false positives and we managed to collect 442 true similes from the web. Manual reviewer had to decide whether they are just comparisons or they contain figure of speech.

| Website | Number of documents |
|---|---|
| Burek.com | 22925 |
| Laguna.rs | 9574 |
| Rastko.rs | 2947 |
| Tarzanija.com | 4792 |
| Total | 40239 |

*Table 1. Number of documents in dataset obtained by crawlers.*

We, also, merged our corpora obtained from the web with the simile corpus published by Vuk Karadžić (Karadžić, 1849). In this publication, Karadžić collected 333 similes. There is relatively small intersection with our corpus of around 10%. Some simile collected by Karadžić were more complex, which could be due to the termination rule on the first noun in our model, so the model might be missing some long and complex similes. Although this is an obvious limitation of our system, for collection of complex similes, we will rely on crown-sourcing interface. We will provide more analysis of similes in Summary and Discussion section.

Our methodology also relies on crowd-sourcing. We provided web interface where users can add missing similes and curator can approve or reject them. At the moment of writing this paper, we had 787 approved similes in our corpus. This number represent the total number of merged simile corpora collected from the world wide web, using our semi-automated text mining methodology, similes collected by Vuk Stefanović Karadžić and similes collected using crowdsourcing.

In order to automate the process further, we experimented with a number of machine learning algorithms to classify true similes from the false positives. We created a dataset with 300 true similes and 300 false positives. These examples were created from the output of our text mining method. The results are presented in *Table 2*: Results of machine learning based classification of similes.

| Algorithm | Precision | Recall | F-Measure |
|---|---|---|---|
| Random Forests | 0.783 | 0.732 | 0.756 |
| Naive Bayes | 0.782 | 0.773 | 0.777 |
| SVM | 0.804 | 0.791 | 0.797 |

*Table 2: Results of machine learning based classification of similes*

The algorithm that perform the best was SVM, with the use of Sequential Minimal Optimization as a solver for quadratic equations (Platt, 1999) and polynomial kernel.

# 5. Summary and Discussion

## 5.1 Summary

In this paper we present a corpus of similes in Serbian language and a methodology for semi-automated creation of simile corpus. Our corpus, with 787 similes is, by the best of our knowledge, the largest simile corpus collected in Serbian language. We believe that, by the mining more web pages for similes and by the use of crowdsourcing, our corpus will grow and remain up to date. Our methodology could be used to model and collect other figures of speech or similes in other languages. The only requirement is to model the particular figure of speech with particular parts of speech.

We published our corpus online[5]. The code of the simile extractor can be also found on GitHub[6].

## 5.2 Analysis of the Corpora

The corpus of similes is able to provide a valuable insight on language development, custom and cultural changes. For example, it could be noted that simile "bori se kao hala s berićetom" ("he fights like a hala with abundance") is nowadays relatively rarely used, especially in urban areas, and it contains elements from national mythology (hala) and from Turkish language (berićet). Hala or ala is a mythical being that brings hail, storm and rain (Petrović, 2000). On the other hand, mythology is used differently in modern similes. For example, in simile "smorio se kao zmaj u vatrogasnoj stanici" ("bored like a dragon at a fire station"), simile is probably no more referring to the dragon (zmaj) from the national mythology, but to the dragon taken from the western mythologies and popularised by Hollywood movies. Dragon (zmaj) in Serbian national mythology is a mythical being that is fighting against the halas and cares about communities' welfare by keeping hails and storms away from their farms (Petrović, 2000). Although, dragon from national mythology is, also a fire creature, fire is not its main characteristic. However, in the simile, a dragon can be very bored in fire station only if it cannot set a fire, which is more likely referring to the western type of dragon, often described as an evil and fearful creature throwing fire. Obviously, there is a shift from the national mythology, since in internet corpus, except the simile mentioning a dragon, we have not encountered any other mention of mythical being, while in Karadžić's corpus there are many, such as "crven kao vampir" ("red like a vampire"), "ždere kao hala" ("eats like a hala"), načinio se kao dodola (he beautified himself like a dodola), etc.

---

[5] http://ezbirka.starisloveni.com
[6] https://github.com/nikolamilosevic86/SerbianComparisonExtractor

On the other hand, in the similes collected from the web, there can be noted smaller amount of Turkish words, which were common in Karadžić's corpus. This might be due to the long Ottoman occupation of Serbia, which was still in place in large parts of country until late 19th century. Examples of similes containing Turkish words are "skupili se kao čifuti u dvoru" ("gathered like Jews in a court", where "čifuti" is a Turkish pejorative name for Jews) or "bled kao akrep" ("pale like a scorpion", where "akrep" is Turkish word for scorpion, however, this word was, also used to describe enormously ugly creature or person).

Compared to Karadžić's corpus of similes, in our corpus obtained from the World Wide Web we could notice the appearance of words, names of things and brands that were not common or did not exist in 19th century. Some examples of such similes are "bankrotiran kao žirant" (bankrupt as an endorser), "čuven kao Lorens od Arabije" (famous like Lorens of Arabia), hitri kao Ferari (quick like a Ferrari), "imati reputaciju kao Hitler" (he has reputation like Hitler), "ima želudac kao Erbas A330" (he has a stomach like an Airbus A330), "jak kao monetarni fond" (strong like a monetary fund), etc.

These are only a few notes about simile corpora. We would like to challenge linguists, culturologists, ethnologists and etymologists to take advantage the simile corpora we provide and analyse it further.

### 5.3 Future Work

As we described in this paper, we created a semi-automated methodology for collecting similes in Serbian language. We also created a website where visitors can add missing similes. However, we would like to see it as a platform for curating short figures of speech in Serbian language. In order to do that, we also provided an interface to add, view and edit proverbs and puzzles. However, these types of figures of speech are currently relying on manual curation. In future we would like to examine possibility of automated or semi-automated curation of other figures of speech, using text mining techniques. As we mentioned before, our text mining method is not able to recognize complex similes containing several nouns correctly. This could, also, be a direction for future research.

### Acknowledgments

We would like to thank Romana Radović on help with relevant literature and her support during the project.